\documentclass[10pt,twocolumn,letterpaper]{article}

\usepackage{wacv}
\usepackage{times}

\usepackage{amsmath}
\usepackage{amssymb}
\usepackage{adjustbox}
\usepackage{booktabs}
\usepackage{caption}
\usepackage{comment}
\usepackage{epsfig}
\usepackage{enumitem}
\usepackage{graphicx}
\usepackage{gensymb}
\usepackage{multirow}
\usepackage{xspace}
\usepackage{xfrac}
\usepackage{subcaption}
\usepackage{float}
\usepackage[percent]{overpic}
\usepackage[colorlinks,pagebackref=true,citecolor=green,bookmarks=false,hypertexnames=true]{hyperref}

\def\wacvPaperID{1013} 

\wacvfinalcopy 

\begin{document}

\title{\LARGE \bf
Generalized Object Detection on Fisheye Cameras for Autonomous Driving: Dataset, Representations and Baseline
}

\author{
Hazem Rashed$^{1}$, 
Eslam Mohamed$^{1}$, 
Ganesh Sistu$^{2}$, \\
Varun Ravi Kumar$^{3}$,
Ciar\'{a}n Eising$^{4}$,
Ahmad El-Sallab$^{1}$ and
Senthil Yogamani$^{2}$ \\ 
$^{1}$Valeo R\&D, Egypt \hspace{0.1cm}
$^{2}$Valeo Vision Systems, Ireland \hspace{0.1cm} \\
$^{3}$Valeo DAR Kronach, Germany \hspace{0.1cm}
$^{4}$University of Limerick, Ireland
}

\makeatletter
\g@addto@macro\@maketitle{
\begin{figure}[H]
  \captionsetup{singlelinecheck=false, font=small, skip=4pt, belowskip=-6pt}
  \setlength{\linewidth}{\textwidth}
  \setlength{\hsize}{\textwidth}
  \vspace{-7mm}
  \centering
  \renewcommand{\tabcolsep}{1pt} 
\begin{tabular}{ccc}
 	\begin{overpic}[width=0.325\textwidth]{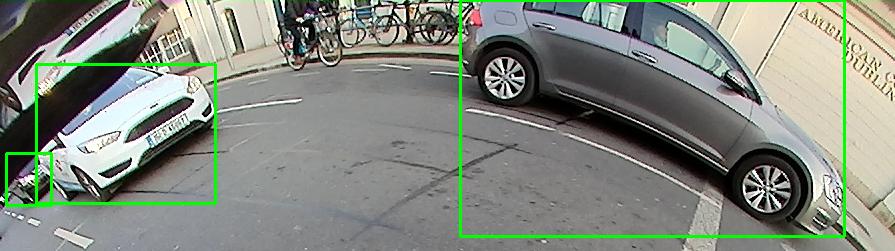}
    \put (0,22) {\colorbox{green}{$\displaystyle\textcolor{black}{\text{(a)}}$}}
    \end{overpic}
    \begin{overpic}[width=0.325\textwidth]{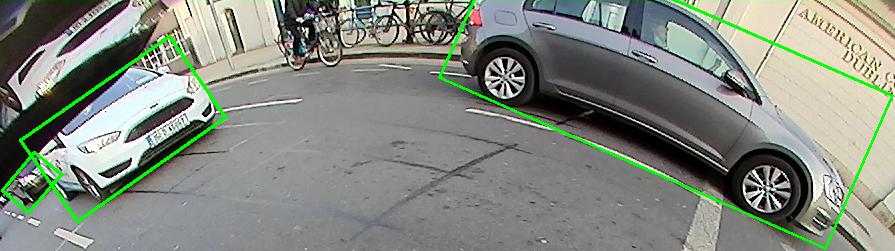}
    \put (0,22) {\colorbox{green}{$\displaystyle\textcolor{black}{\text{(b)}}$}}
    \end{overpic}
    \begin{overpic}[width=0.325\textwidth]{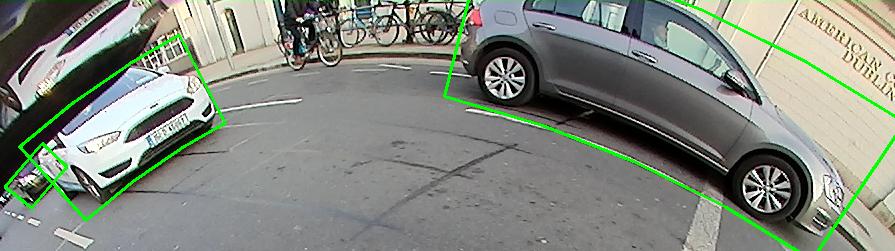}
    \put (0,22) {\colorbox{green}{$\displaystyle\textcolor{black}{\text{(c)}}$}}
    \end{overpic} \\

 	\begin{overpic}[width=0.325\textwidth]{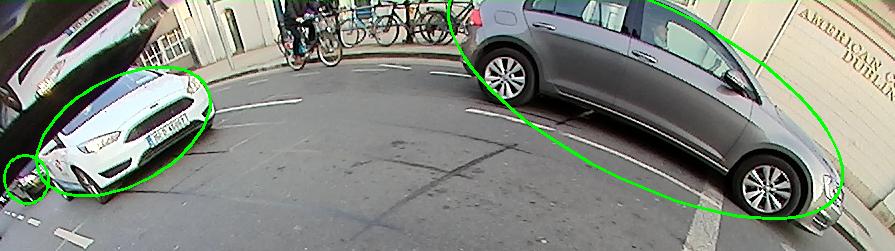}
    \put (0,22) {\colorbox{green}{$\displaystyle\textcolor{black}{\text{(d)}}$}}
    \end{overpic} 
    \begin{overpic}[width=0.325\textwidth]{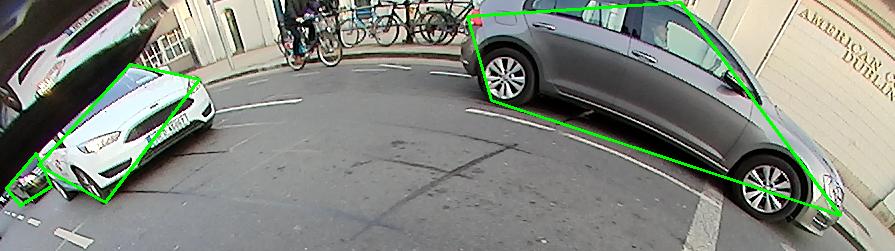}
    \put (0,22) {\colorbox{green}{$\displaystyle\textcolor{black}{\text{(e)}}$}}
    \end{overpic} 
    \begin{overpic}[width=0.325\textwidth]{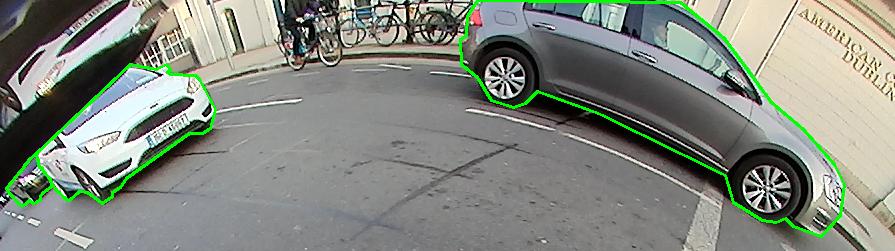}
    \put (0,22) {\colorbox{green}{$\displaystyle\textcolor{black}{\text{(f)}}$}}
    \end{overpic}
\end{tabular}
\vspace{-2mm}
\caption{\textbf{Various 2D object detection representations on fisheye camera images.} (a) Standard Box, (b) Oriented Box, (c) Curved Box, (d) Ellipse, (e) 4-sided Polygon and (f) 24-sided Polygon.}
\label{fig:abstract}
\end{figure}
\vspace{5mm}
}
\makeatother
\maketitle
\thispagestyle{empty}
\begin{abstract}
Object detection is a comprehensively studied problem in autonomous driving. However, it has been relatively less explored in the case of fisheye cameras. The standard bounding box fails in fisheye cameras due to the strong radial distortion, particularly in the image's periphery. We explore better representations like oriented bounding box, ellipse, and generic polygon for object detection in fisheye images in this work. We use the IoU metric to compare these representations using accurate instance segmentation ground truth. We design a novel curved bounding box model that has optimal properties for fisheye distortion models. We also design a curvature adaptive perimeter sampling method for obtaining polygon vertices, improving relative mAP score by 4.9\% compared to uniform sampling.  Overall, the proposed polygon model improves mIoU relative accuracy by 40.3\%. It is the first detailed study on object detection on fisheye cameras for autonomous driving scenarios to the best of our knowledge. The dataset\footnote{This dataset is an extension of our WoodScape dataset \cite{woodscape}.} comprising of 10,000 images along with all the object representations ground truth will be made public to encourage further research. We summarize our work in a short video with qualitative results at {\url{https://youtu.be/iLkOzvJpL-A}}.
\end{abstract}
\section{\textit{Introduction}}

\begin{figure}[t]
\centering
  \captionsetup{singlelinecheck=false, font=small, skip=4pt, belowskip=-12pt}
    \includegraphics[width=0.95\linewidth]{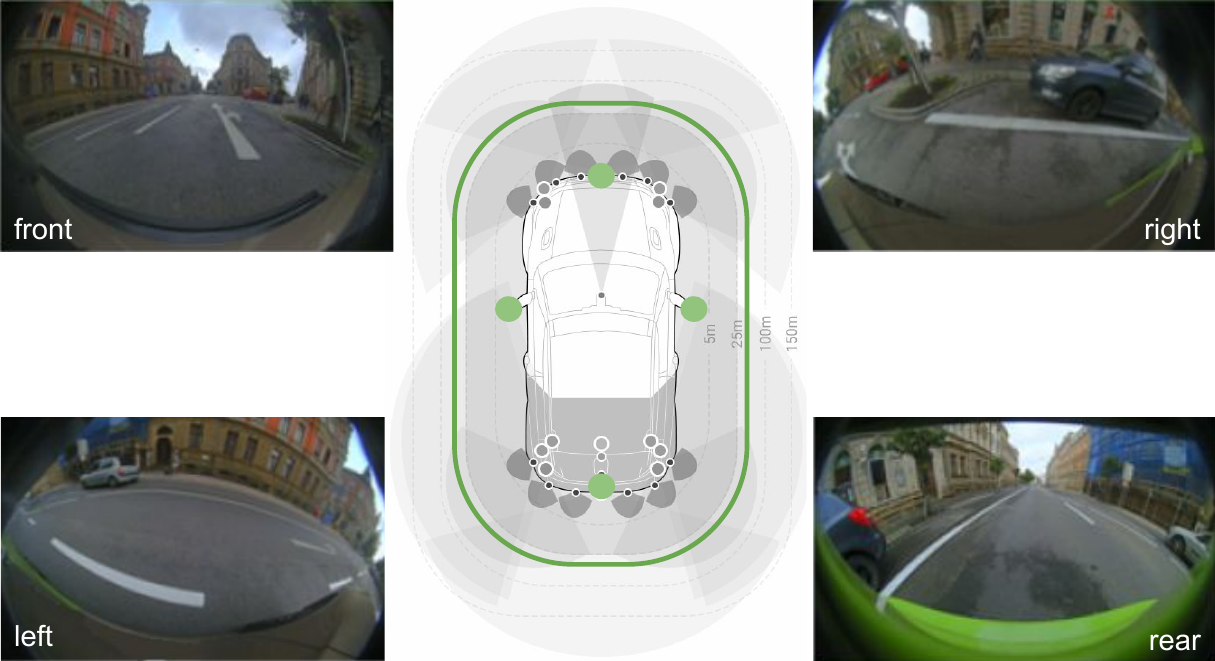}
  \caption{\textbf{Illustration of four fisheye camera images} (locations in the car are marked with green circles) forming the surround-view camera network. Wide field of view of 190$^{\circ}$ and strong radial distortion can be observed.}
\label{fig:car_new}
\end{figure}

Typically, automotive systems are equipped with a multi-camera network to cover all the field of view and range \cite{horgan2015vision}. Four surround-view fisheye cameras are typically part of this sensor suite, as illustrated in Figure \ref{fig:car_new}. We can accomplish a dense $360\degree$ near field perception with the employment of four surround-view fisheye cameras, making them suitable for automated parking, low-speed maneuvering, and emergency braking \cite{heimberger2017computer}. The wide field of view of the fisheye image comes with the side effect of strong radial distortion. Objects at different angles from the optical axis look quite different, making the object detection task a challenge. A common practice is to rectify distortions in the image using a fourth order polynomial \cite{woodscape} model or unified camera model \cite{khomutenko2015enhanced}. However, undistortion comes with resampling distortion artifacts, especially at the periphery. In particular, the negative impact on computer vision due to the introduction of spurious frequency components is understood \cite{LourencoSIFT}. In addition, other more minor impacts include reduced field of view and non-rectangular image due to invalid pixels. Although semantic segmentation is an easier solution on fisheye images, object detection annotation costs are much lower \cite{siam2017deep}. In general, there is limited work on fisheye perception \cite{kumar2018monocular,yahiaoui2019fisheyemodnet,uvrivcavr2019soilingnet,kumar2020fisheyedistancenet,kumar2020unrectdepthnet,kumar2020syndistnet}.

We can broadly classify the state-of-the-art object detection methods based on deep learning into two types: two-stage detectors and single-stage detectors. Agarwal~\etal~\cite{agarwal2018recent} provides a detailed survey of current object detection methods and its challenges. Relatively, fisheye camera object detection is a much harder problem. The rectangular bounding box fails to be a good representation due to the massive distortion in the scene. As demonstrated in Figure~\ref{fig:abstract} (a), the size of the standard bounding box is almost double the size of the object of interest inside it. Instance segmentation can help obtain accurate object contours. However, it is a different task which is computationally complex and more expensive to annotate. It also typically needs a bounding box estimation step. There is relatively less work on object detection for fisheye or closely related omnidirectional cameras. One of the main issues is the lack of a useful dataset, particularly for autonomous driving scenarios. The recent fisheye object detection paper FisheyeDet~\cite{li2020fisheyedet} emphasizes the lack of a useful dataset, and they create a simulated fisheye dataset by applying distortions to the Pascal VOC dataset \cite{everingham2010pascal}. FisheyeDet makes use of a 4-sided polygon representation aided by distortion shape matching. SphereNet~\cite{coors2018spherenet} and its variants \cite{perraudin2019deepsphere, su2019kernel, jiang2019spherical} formulate CNNs on spherical surfaces. However, fisheye images do not follow spherical projection models, as seen by non-uniform distortion in horizontal and vertical directions.\par
Our objective is to present a more detailed study of various techniques for fisheye object detection in autonomous driving scenes. Our main contributions include:
\begin{itemize}[nosep]
    \item Exploration of seven different object representations for fisheye object detection. 
    \item Design of novel representations for fisheye images, including the curved box and adaptive step polygon.
    \item Public release of a dataset of 10,000 images with annotations for all the object representations.
    \item Implementation and empirical study of  FisheyeYOLO baseline, which can output different representations.
\end{itemize}
\section{\textit{Object Representations}} 
\label{sec:representations}

\subsection{Adaptation of Box representations}

\textbf{\textit{Standard Box Representation}}
The rectangular bounding box is the most common representation for object detection. They are aligned to the pixel grid axes, which makes them efficient to be regressed using a machine learning model. They are represented by four parameters ($\hat{x}$, $\hat{y}$, $\hat{w}$, $\hat{h}$), namely the box center, width and height. It has the advantage of simplified, low-cost annotation. It works in most cases, but it may capture a large non-object area within the box for complex shapes. It is particularly the case for fisheye distorted images, as shown in Figure \ref{fig:abstract} (a).\par
\textbf{\textit{Oriented Box Representation}}
The oriented box is a simple extension of the standard box with an additional parameter $\hat{\theta}$ to capture the rotation angle of the box. It is also referred to as a titled or rotated box. Lienhart et al. \cite{lienhart2002extended} adapted Viola-Jones object detection framework to output rotated boxes. It is also commonly used in lidar top-view object detection methods \cite{Geiger2012CVPR}. The orientation ground-truth range spans the range of (-90\degree to +90\degree) where this rotation angle is defined with respect to the x-axis. For this study, we used instance segmentation contours to estimate the optimally oriented box as a minimum enclosing rectangle.\par
\textbf{\textit{Ellipse Representation}}
Ellipse is closely related to an oriented box and can be represented using the same parameter set. Width and height parameters represent the major and minor axis of the ellipse. In contrast to an oriented box, the ellipse has a smaller area at the edge, and thus it is better for representing overlapping objects as shown for the objects at the left end in Figure \ref{fig:abstract}. It may also help fit some objects like vehicles better than a box.  We created our ground truth by fitting a minimum enclosing ellipse to the ground truth instance segmentation contours. In parallel work, Ellipse R-CNN \cite{dong2020ellipse} used ellipse representation for objects instead of boxes.\par
\subsection{Distortion aware representation} \label{sec:curvedbox}

This subsection aims to derive an optimal representation of objects undergoing radial distortion in fisheye images assuming a rectangular box is optimal for pinhole cameras. In the pinhole camera with no distortion, a straight line in the scene is imaged as a straight line in the image. A straight line in the scene is imaged as a curved segment in the image for a fisheye image. The specific type of fisheye distortion determines the nature of the curved segment. The fisheye cameras from the dataset we used are well represented and calibrated using a 4\textsuperscript{th} order polynomial model for the fisheye distortion \cite{woodscape}. The author's are aware that there have been many developments in fisheye camera models over the past few decades, e.g. \cite{kannala2006fisheye, brauerDivisionModel, Khomutenko2016eucm}. In this section, we consider the fourth order polynomial model and the division model only. The reason is that the fourth order polynomial model is provided by the data set that we use, and we examine the division model to understand if the use of circular arcs is valid under such fisheye projections.

In this case, the projection of a line on to the image can be described parametrically with complicated polynomial curves. Let us consider a much simpler model for the moment - a first-order polynomial (or equidistant) model of a fisheye camera. \ie $r' = a\theta$, where $r'$ is the radius on the image plane, and $\theta$ is the angle of the incident ray against the optical axis. If we consider the parametric equation $\mathbf{P}(t)$ of a line in 3D Euclidean space:
\begin{equation}
    \mathbf{P}(t) = \mathbf{D}t + \mathbf{Q}
\end{equation}
where $\mathbf{D} = [D_x, D_y, D_z]$ is the direction vector of the line and $\mathbf{Q} = [Q_x, Q_y, Q_z]$ is a point through which the line passes. Hughes et al. \cite{hughesFisheye} have shown that the projection on to a fisheye camera that adheres to equidistant distortion is described by:
\begin{equation}
    \mathbf{p}'(t) = \left[\begin{matrix}
        D_x t + Q_x \\
        D_y t + Q_y
    \end{matrix}\right]
    \frac{|\mathbf{p}'(t)|}{|\mathbf{p}(t)|}
\end{equation}
where
\begin{align}
    \frac{|\mathbf{p}'(t)|}{|\mathbf{p}(t)|} &=
    \frac{a \arctan{\left(\frac{d_{xy}(t)}{D_z t + Q_z}\right)}}{d_{xy}(t)} \\
    d_{xy}(t) &= \sqrt{(D_x t + Q_x)^2 + (D_y t + Q_y)^2}
\end{align}
$\mathbf{p}(t)$ is the projected line in a pinhole camera, and $\mathbf{p}'(t)$ is the distorted image of the line in a fisheye camera.
\begin{figure}
     \centering
     \begin{subfigure}[b]{0.23\textwidth}
         \centering
         \includegraphics[width=\textwidth]{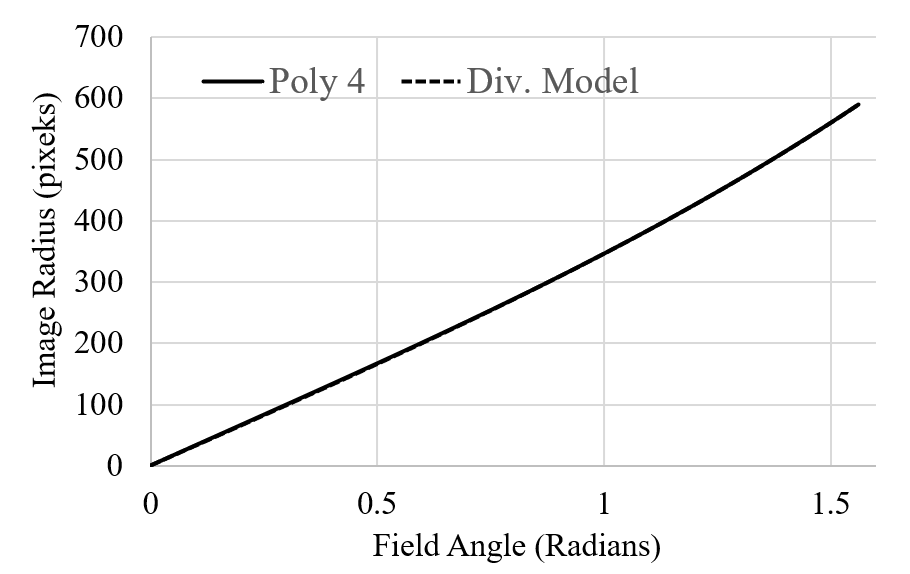}
         \caption{}
     \end{subfigure}
     \begin{subfigure}[b]{0.23\textwidth}
         \centering
         \includegraphics[width=\textwidth]{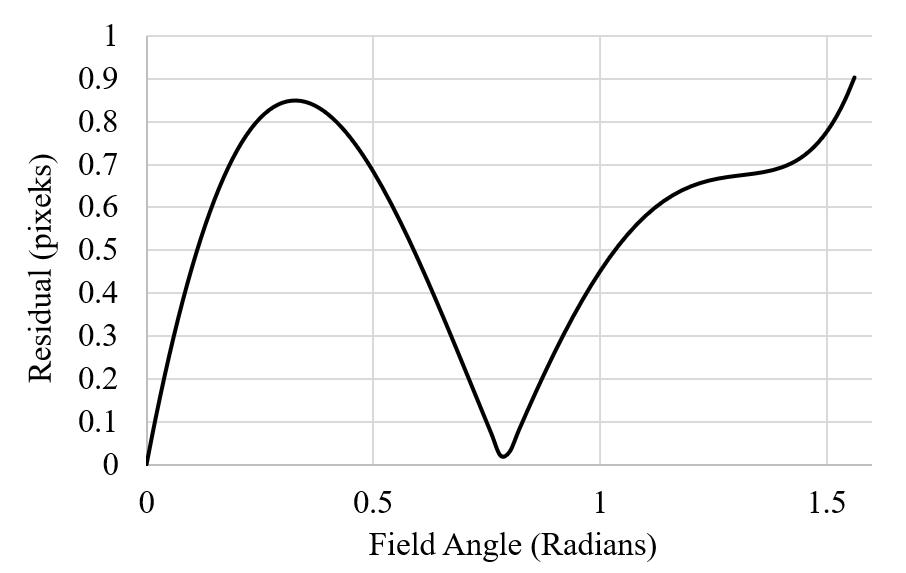}
         \centering\caption{}
     \end{subfigure}
  \captionsetup{singlelinecheck=false, font=small, skip=4pt, belowskip=-6pt}
\caption{\textbf{Approximation of 4\textsuperscript{th} order radial distortion model by division model.} (a) shows the division model fit to the 4\textsuperscript{th} order polynomial model. Note that the two are almost indistinguishable. (b) shows the residual error per field angle.}
\label{fig:divmodel}
\end{figure}

This is a complex description of a straight line's projection, especially considering we have ignored all but the first-order polynomial term. Therefore, it is highly desirable to describe straight lines' projection using a more straightforward geometric shape.
\begin{figure}[t]
  \captionsetup{singlelinecheck=false, font=small, skip=4pt, belowskip=-10pt}
   \centering
   \includegraphics[width=0.4\textwidth]{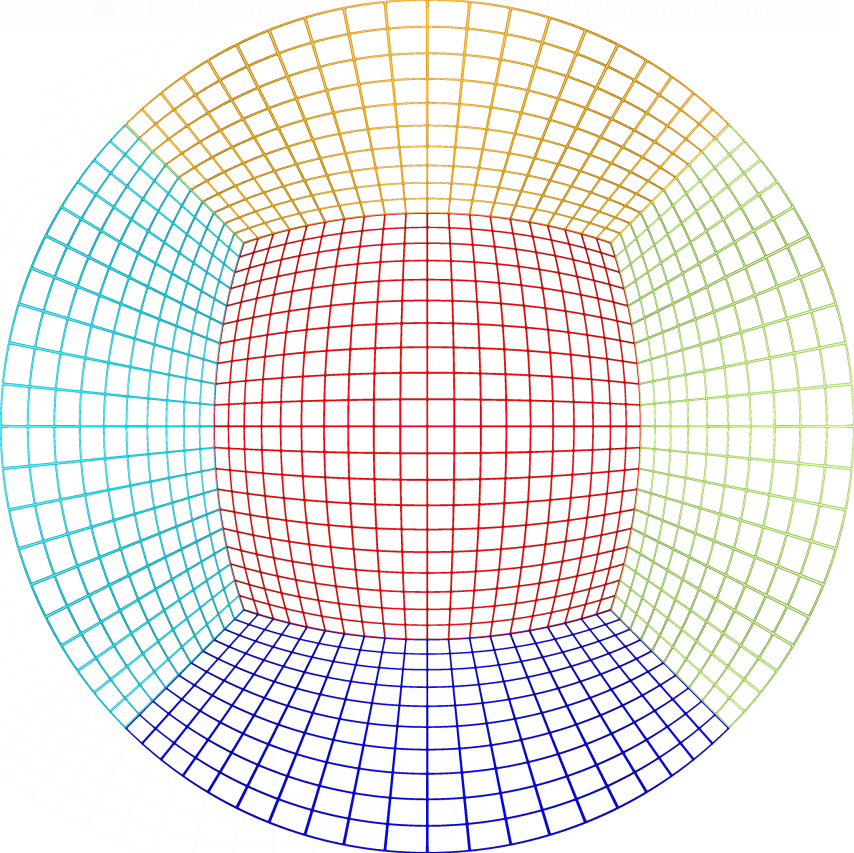}
   \vspace{10pt}
   \includegraphics[width=0.48\textwidth]{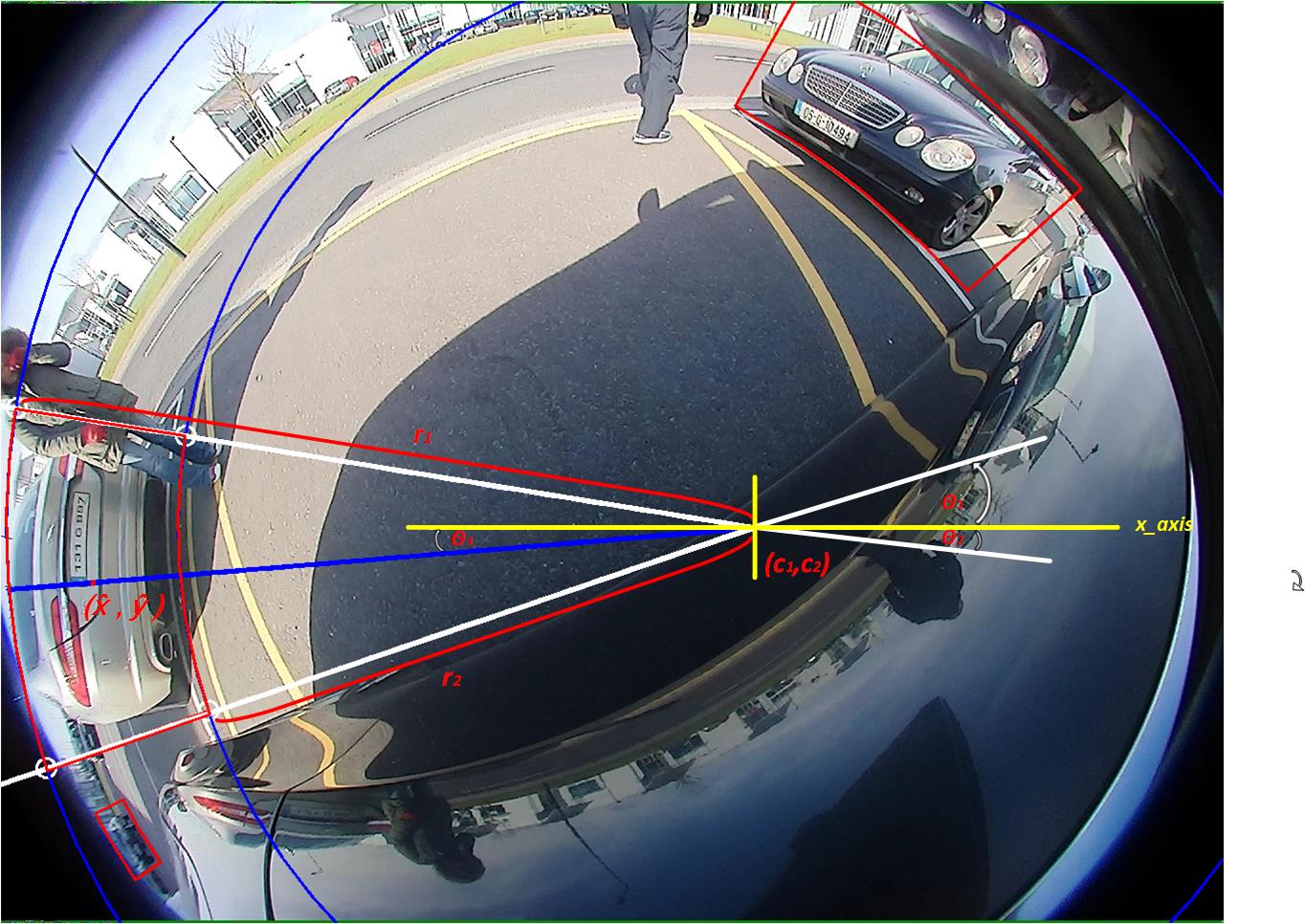}
   \caption{\textbf{Top:} Illustration of fisheye distortion of projection of an open cube. A 4\textsuperscript{th}-degree polynomial model radial distortion. We can visually notice that box matures to a curved box, and it is justified theoretically in Section~\ref{sec:curvedbox}. \textbf{Bottom:} We propose the \textbf{Curved Bounding Box} using a circle with an arbitrary center and radius, as illustrated. It captures the radial distortion and obtains a better footpoint. The center of the circle can be equivalently reparameterized using the object center ($\hat{x}$, $\hat{y}$).}
   \label{fig:curved_box_samples}
\end{figure}
\begin{figure*}[t]
\captionsetup{singlelinecheck=false, font=small,  belowskip=-12pt}
\centering
    \includegraphics[width=0.99\linewidth]{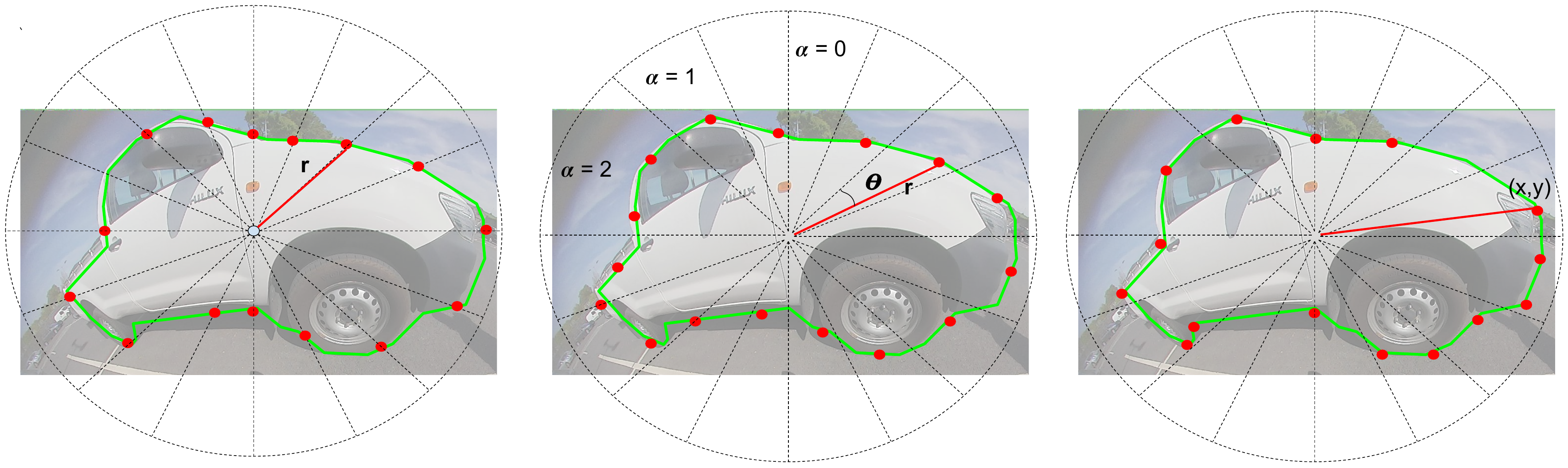}
   \caption{\textbf{Generic Polygon Representations.} \textbf{Left:} Uniform angular sampling where the intersection of the polygon with the radial line is represented by one parameter per point (r). \textbf{Middle:} Uniform contour sampling using L2 distance. It can be parameterized in polar co-ordinates using 3 parameters ($r$, $\theta$, $\alpha$). $\alpha$ denotes the number of polygon vertices within the sector, and it may be used to simplify the training. Alternatively, 2 parameters (x,y) can be used, as shown in the figure on the right. \textbf{Right:} Variable step contour sampling. It is shown that the straight line in the bottom has less number of points than curved points such as the wheel. This representation allows to maximize the utilization of vertices according to local curvature.}
\label{fig:polygon_representation}
\end{figure*}
Bräuer-Burchardt and Voss \cite{brauerDivisionModel} show that if the first-order \textit{division model} can accurately describe the fisheye distortion, then we may use circles in the image to model the projected straight lines. As a note, the division model is generalised in \cite{scaramuzzaFisheye}, though it loses the property of straight line to circular arc projection. We should then consider how well the division model fits with the 4\textsuperscript{th} order polynomial model. In~\cite{hughesFisheye}, the authors adapt the division model slightly to include an additional scaling factor and prove that this does not impact the projection of line to a circle. They show that the division model is a correct replacement for the equidistant fisheye model. Here we repeat this test but compare the division model to the 4\textsuperscript{th} order polynomial. The results are shown in Figure \ref{fig:divmodel}. As can be seen, the division model can map to the 4\textsuperscript{th} order polynomial with a maximum of less than 1-pixel error. While this may not be accurate enough for applications in which sub-pixel error is desirable, it is sufficient for bounding box accuracy.\par

Therefore, we propose a novel curved bounding box representation using circular arcs. Figure~\ref{fig:curved_box_samples} (top) provides a visual justification of circular arcs. We illustrate an open cube's projection with grid lines where the straight lines become circular arcs after projection. Figure~\ref{fig:curved_box_samples} (bottom) illustrates the details of the curved bounding box. The blue line represents the axis, and the white lines intersect with the circles creating starting and ending points of the polygon. This representation allows two sides of the box to be curved, giving the flexibility to adapt to image distortion in fisheye cameras. It can also specialize in an oriented bounding box when there is no distortion for the objects near the principal point.\par

We create an automatic process to generate the representation that takes an object contour as an input. First, we generate an oriented box from the output contour. We choose a point that lies on the oriented box's axis line to represent a circle center. From the center, we create two circles intersecting with the corner points of the bounding box. We construct the polygon based on the two circles and the intersection points. To find the best circle center, we iterate over the axis line and choose the circle center, which forms a polygon with the minimum IoU with the instance mask. The output polygon can be represented by 6 parameters, namely, ($c_{1}$, $c_{2}$, $r_{1}$, $r_{2}$, $\theta_{1}$, $\theta_{2}$) representing the circle center, two radii and angles of the start and end points of the polygon relative to the horizontal x-axis. By simple algebraic manipulation, we can re-parameterize the curved box using the object center ($\hat{x}$, $\hat{y}$) following a typical box representation instead of the center of the circle.\par
\subsection{Generic Polygon Representations}

Polygon is a generic representation for any arbitrary shape and is typically used even, for instance segmentation annotation. Thus polygon output can be seen as a coarse segmentation. We discuss two standard representations of a polygon and propose a novel extension that improves accuracy.\par
\textbf{\textit{Uniform Angular Sampling}} 
Our polar representation is quite similar to PolarMask~\cite{polarmask} and PolyYOLO~\cite{polyyolo} approaches. As illustrated in Figure~\ref{fig:polygon_representation} (left), the full angle range of $360\degree$ is split into $N$ equal parts where $N$ is the number of polygon vertices. Each polygon vertex is represented by the radial distance $r$ from the centroid of the object. Uniform angular sampling removes the need for encoding $\theta$ parameter. Polygon is finally represented by object center ($\hat{x}$, $\hat{y}$) and  \{$r_{i}$\}.\par
\textbf{\textit{Uniform Perimeter Sampling}} 
In this representation, we divide the perimeter of the object contour equally to create $N$ vertices. Thus the polygon is represented by a set of vertices \{($x_{i}$, $y_{i}$)\} using the centroid of the object as the origin. PolyYOLO~\cite{polyyolo} showed that it is better to learn polar representation of the vertices \{($r_{i}$, $\theta_{i}$)\} instead. They define a parameter $\alpha$ to denote the presence or absence of a vertex in a sector, as shown in Figure~\ref{fig:polygon_representation} (middle). We extend this parameter to be the count of vertices in the sector.\par
\textbf{\textit{Curvature-adaptive Perimeter Sampling}}  
The original curve in the object contour between two vertices gets approximated by a straight line in the polygon. For regions of high curvature, this is not a good approximation. Thus, we propose an adaptive sampling based on the curvature of the local contour. We distribute the vertices non-uniformly in order to represent the object contour best. Figure~\ref{fig:polygon_representation} (right) shows the effectiveness of this approach, where a larger number of vertices are used for higher curvature regions than straight lines, which can be represented by lesser vertices.  We adopt the algorithm in~\cite{teh1989detection} to detect the dominant points in a given curved shape, which best represents the object. Then we reduce the set of points using the algorithm in~\cite{douglas1973algorithms} to get the most representative simplified curves. This way, our polygon has dense points on the curved parts and sparse points on the straight parts, which maximize the utilization of the predefined number of points per contour. 
\section{FisheyeYOLO network}
\begin{figure}[t]
  \captionsetup{singlelinecheck=false, font=small, skip=4pt, belowskip=-10pt}
  \centering
  \includegraphics[width=0.5\textwidth]{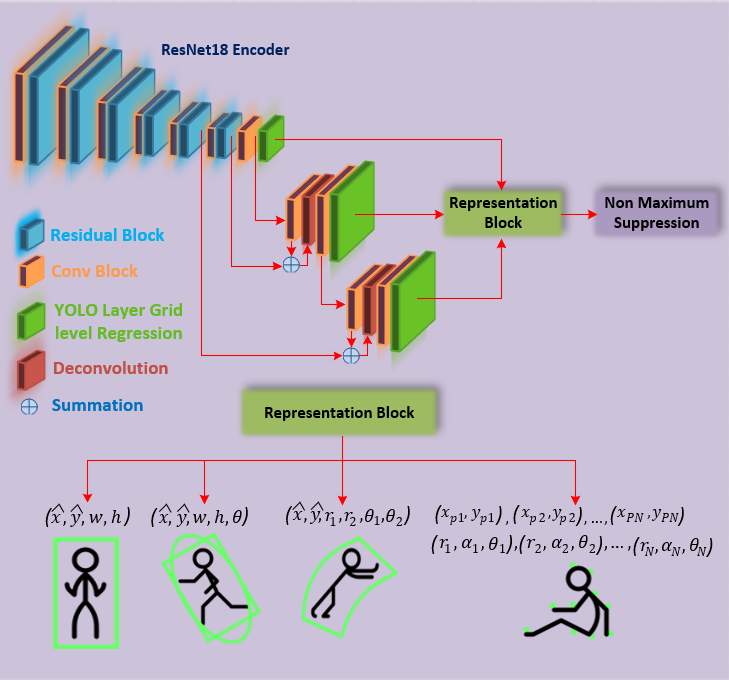}
  \caption{\textbf{FisheyeYOLO is an extension of YOLOv3} which can output different output representations discussed in Section \ref{sec:representations}.}
\label{fig:generalized_yolo}
\end{figure}

We adapt YOLOv3 \cite{YOLOV3} model to output different representations discussed in Section \ref{sec:representations}. We call this FisheyeYOLO, as illustrated in Figure \ref{fig:generalized_yolo}. Our baseline bounding box model is the same as YOLOv3 \cite{YOLOV3}, except the Darknet53 encoder is replaced with ResNet18 encoder. Similar to YOLOv3, object detection is performed at multiple scales. For each grid in each scale, object width ($\hat{w}$), height ($\hat{h}$), object center coordinates ($\hat{x}$, $\hat{y}$) and object class is inferred. Finally, a non-maximum suppression is used to filter out the low confidence detections. Instead of using $L_{2}$ loss for categorical and objectness classification, we used standard categorical cross-entropy and binary entropy losses, respectively. The final loss is a combination of sub-losses, as illustrated below:
\begingroup\makeatletter\def\f@size{8}\check@mathfonts
\begin{align}
\mathcal{L}_{xy} &= \lambda_{coord} \sum_{i=0}^{S^2}\sum_{j=0}^{B} l_{ij}^{obj}[(x_{i}-\hat{x}_{_{i}})^2 + (y_{i}-\hat{y}_{_{i}})^2]\\
\mathcal{L}_{wh} &= \lambda_{coord} \sum_{i=0}^{S^2}\sum_{j=0}^{B} l_{ij}^{obj}[(\sqrt{w_{i}}- \sqrt{\hat{w}_{_{i}}})^2 + (\sqrt{h_{i}}-\sqrt{\hat{h}_{_{i}}})^2] \\   
\mathcal{L}_{obj} &= -\sum_{i=0}^{S^2}\sum_{j=0}^{B} [C_{i}log(\hat{C}_{_{i}})]\\
\mathcal{L}_{class} &=  -\sum_{i=0}^{S^2} l_{ij}^{obj}\sum_{c=\text{classes}}[c_{i,j}log(p({\hat{c}_{_{i,j}}}))] \\
\mathcal{L}_{total} &= \mathcal{L}_{xy} + \mathcal{L}_{wh} + \mathcal{L}_{obj} + \mathcal{L}_{class}
\end{align}
\endgroup
where height and width are predicted as offsets from pre-computed anchor boxes.

In the case of oriented box or ellipse prediction, we define an additional loss function based on ellipse angle or orientation of the box. The loss function for oriented box and ellipse is:
\begin{align}
\mathcal{L}_{orn} &= \sum_{i=0}^{S^2}\sum_{j=0}^{B} l_{ij}^{obj}[\theta_{i}-\hat{\theta}_{_{i}}]^2 \\
\mathcal{L}_{total} &= \mathcal{L}_{xy} + \mathcal{L}_{wh} + \mathcal{L}_{obj} + \mathcal{L}_{class} + \mathbf{\mathcal{L}_{orn}}\label{eq:orn_loss}
\end{align}
where $\mathcal{L}_{total}$, is the total loss minimized for oriented box regression. In case of curved box, $\mathcal{L}_{wh}$ is replaced by $\mathcal{L}_{cods}$ in equation \eqref{eq:rtheta_loss}.\par
We also explored methods of learning orientation as a classification problem instead of a regression problem. One motivation is due to discontinuity of angles at $90\degree$ due to wrapping around of angles.
In this scenario, we discretized the orientation into 18 bins, where each bin represents a range of 10\degree with a tolerance of +-5\degree. To further improve our prediction, we design an IoU loss function that guides the model to minimize the difference in the area of the predicted box and the ground truth box. We compute the area of the predicted and ground truth rectangles and apply regression loss on those values. This loss maximizes the overlapping area between the prediction and the ground truth by improving the overall results. The IoU loss is,
\begin{align}
\mathcal{L}_{IoU} &= \lambda_{coord} \sum_{i=0}^{S^2}\sum_{j=0}^{B} l_{ij}^{obj}[(a_{i}-\hat{a}_{_{i}})^2]
\end{align}
where $a$ represents the area of the representation at hand. We report all the results related to these experiments in Table \ref{tab:ablation}.\par

The polar polygon regression loss is,
\begin{align}
  \mathcal{L}_{cods} &= \sum_{i=0}^{S^2} \sum_{j=0}^{N} \hat{\alpha}_{ij} [(r_{i,j}-\hat{r}_{_{i,j}})^2 + (\theta_{i,j}-\hat{\theta}_{_{i,j}})^2] \label{eq:rtheta_loss}\\
  \mathcal{L}_{mask} &=-\sum_{i=0}^{S^2} \sum_{j=0}^{N} \alpha_{ij}
log(\hat{\alpha}_{ij}) \\
  \mathcal{L}_{total} &= \mathcal{L}_{xy} + \mathcal{L}_{obj} + \mathcal{L}_{class} + \mathbf{\mathcal{L}_{cods}} + \mathbf{\mathcal{L}_{mask}}
\end{align}
where N corresponds to the number of sampling points, each point is sampled with a step size of $360/N$ angle in polar coordinates, as shown in Figure \ref{fig:polygon_representation}. Our polar loss is similar to PolyYOLO \cite{polyyolo}, where each polygon point is (in red) is represented using three parameters $r$, $\theta$, and $\alpha$. Hence the total required parameters for $N$ sampling points are $3\times N$. The same is presented in Figure \ref{fig:polygon_representation} (middle). \par
In Cartesian representation, we regress over two parameters ($\hat{x}$, $\hat{y}$) for each polygon point. We further improve our predictions by adding our IoU loss function, which minimizes the area between the prediction and ground truth. We refer to both loss functions as localization loss $\mathcal{L}_{Localization}$. Our combined loss for Cartesian polygon predictions is:
\begin{equation}
\label{insta_yolo_loss}
\mathcal{L}_{total} = \mathcal{L}_{Class} +\mathcal{L}_{Obj} +\mathcal{L}_{Localization}
\end{equation}
where $\mathcal{L}_{Obj}$ and $\mathcal{L}_{Class}$ are inherited from YoloV3 loss functions. According to the representation at hand, we perform the non-maximum suppression. We generate the predictions for all the representations; filter out the low confidence objects—computation of IoU of the output polygon with the list of outputs where high-IoU objects are filtered out.

\begin{table}[t]
\captionsetup{belowskip=-10pt, skip=4pt, font=small, singlelinecheck=false}
\centering
\begin{adjustbox}{width=0.8\columnwidth}
\begin{tabular}{@{}l|c|c|c|c|c|c@{}}
\toprule
\textit{\begin{tabular}[c]{@{}l@{}} \# Vertices\end{tabular}} &
  \begin{tabular}[c]{@{}c@{}}4 \end{tabular} &
  \begin{tabular}[c]{@{}c@{}}12\end{tabular} &
  \begin{tabular}[c]{@{}c@{}}24\end{tabular} &
  \begin{tabular}[c]{@{}c@{}}36\end{tabular} &
  \begin{tabular}[c]{@{}c@{}}60\end{tabular} &
  \begin{tabular}[c]{@{}c@{}}120\end{tabular} \\\hline
mIoU &  85 & 85.3 & 86.6 & 91.8 & 94.2 & 98.4 \\ \bottomrule
\end{tabular}
\end{adjustbox}
\caption{\textbf{Analysis of the number of polygon vertices for representing the objects contour.} mIoU is calculated between the approximated polygon and ground truth instance segmentation mask.}
\label{tab:number_points_analysis}
\end{table}

\section{Experimental Results}

\subsection{Dataset and Evaluation Metrics}
\label{sec:metrics}

Our dataset comprises of 10,000 images sampled roughly equally from the four views. The dataset comprises 4 classes, namely vehicles, pedestrians, bicyclists, and motorcyclists. Vehicles further have sub-classes, namely cars and large vehicles (trucks/buses). The images are in RGB format with 1MPx resolution and $190\degree$ horizontal FOV. The dataset is captured in several European countries and the USA. For our experiments, we used only the vehicles' class. We divide our dataset into {60-10-30} split and train all the models using the same setting. More details are discussed in our \textit{WoodScape Dataset} paper \cite{woodscape}.\par

The objective of this work is to study various representations of fisheye object detection. Conventional object detection algorithms evaluate their predictions against their ground-truth, which is usually a bounding box. Unlike conventional evaluation, our first objective is to provide better representation than a conventional bounding box. Therefore, we first evaluate our representations against the most accurate representation of the object, the ground-truth instance segmentation mask. We report mIoU between a representation and the ground-truth instance mask. 

Additionally,  we qualitatively evaluate the representations in obtaining object intersection with the ground (footpoint). This is critical as it helps localize the object in the map and provide more accurate vehicle trajectory planning. Finally, we report model speed in terms of frames-per-second (fps) as we focus on real-time performance. The distortion is higher in side cameras compared to front and rear cameras. Thus, we provide our evaluation on each camera separately. To simplify our baseline, we only evaluate on vehicles class although four classes are available in the dataset.

\begin{table}[t]
\captionsetup{belowskip=-12pt, skip=4pt, font=small, singlelinecheck=false}
\centering
\begin{adjustbox}{width=\columnwidth}
\begin{tabular}{lcccccc}
\toprule
\multicolumn{1}{l|}{\textit{\textbf{Representation}}} & \multicolumn{4}{c|}{\textbf{mIoU}} & \multicolumn{1}{l|}{\textbf{mIoU}} & 
\textit{\begin{tabular}[c]{@{}l@{}} \textbf{No. of} \\ \textbf{params} \end{tabular}} \\ 
\toprule
\multicolumn{1}{l|}{} & \multicolumn{1}{c|}{Front} & \multicolumn{1}{c|}{Rear} & \multicolumn{1}{c|}{Left} & \multicolumn{1}{c|}{Right} & & \\ 
\midrule

Standard Box  & 53.7 & 47.9 & 60.6 & 43.2 & 51.35 & 4 \\
Curved Box    & 53.7 & 48.6 & 63.5 & 44.2 & 52.5 & 6\\
Oriented Box  & 55 & 50.2 & 64.8 & 45.9 & 53.9 & 5 \\
Ellipse       & 56.5 & 51.7	& 66.5 & 47.5 &	55.5 & 5\\
\hline
4-sided Polygon (uniform) & 70.7 & 70.6 & 70.2 & 69.6 &	70.2 & 8 \\
24-sided Polygon (uniform) & 85 & 84.9 & 83.9 & 83.8 & 84.4 & 48 \\
24-sided Polygon (adaptive) & \textbf{87.2} & \textbf{87} & \textbf{86.2} & \textbf{86.1} & \textbf{86.6} & 48 \\

\bottomrule
\end{tabular}
\end{adjustbox}
\caption{\textbf{Evaluation of representation capacity of various representations.} We estimate the best fit for each representation using ground truth instance segmentation and then compute mIoU to evaluate capacity. We also list the number of parameters used for each representation to provide comparison of complexity. }
\label{tab:gt_table}
\end{table}

\subsection{Results Analysis}

\subsubsection{Number of Polygon Points}

Polygon is a more generic representation of complex object shapes that arise in fisheye images. We perform a study to understand the effect of the number of vertices parameter in a polygon. We use a uniform perimeter sampling method to vary the number of vertices and compare the IoU using instance segmentation as ground truth. The results are tabulated in Table~\ref{tab:number_points_analysis}. A 24-sided polygon seems to provide a reasonable trade-off between the number of parameters and accuracy. Although a 120-sided polygon provides 8\% higher ground truth, it will be difficult to learn this representation and it will produce noisy overfitting. For the quantitative experiments, we fix the number of vertices to be 24 to represent each object. We observe no significant difference in fps due to increasing the number of vertices where our models run at 56 fps on a standard NVIDIA TitanX GPU. It is due to the utilization of YoloV3~\cite{YOLOV3} architecture, which performs the prediction at each grid cell in a parallel manner.\par


\subsubsection{Evaluation of Representation Capacity}

Table \ref{tab:gt_table} compares the performance of different representations using its ground truth fit relative to instance segmentation ground truth. This empirical metric is used to demonstrate the maximum performance a representation can achieve regardless of the model complexity. As expected, a 24-sided polygon achieves the highest mIou showing that it has the best representation capacity. Our proposed curvature-adaptive polygon achieves a {2.2\%} improvement over uniform sampling polygon with the same vertices. Polygon annotation is relatively more expensive to collect, and it increases model complexity. Thus it is still interesting to consider more simpler bounding box representations. 


\begin{table}[!t]
\centering
\captionsetup{belowskip=-10pt, skip=4pt, font= small, singlelinecheck=false}
\scalebox{0.8}{
\begin{adjustbox}{width=0.7\columnwidth}
\begin{tabular}{@{}lc@{}}
\toprule
\textit{\textbf{Representation}}  & 
\textbf{mAP} \\ \midrule
\multicolumn{2}{c}{Oriented Box} \\
\midrule
Orientation regression                & 39             \\
Orientation classification            & 40.6             \\
Orientation classification + IoU loss & \textbf{41.9}             \\
\midrule
\multicolumn{2}{c}{24-sided Polygon} \\
\midrule
Uniform Angular & 55.6 \\
Uniform Perimeter & 55.4 \\
Adaptive Perimeter   & \textbf{58.1} \\ \bottomrule
\end{tabular}
\end{adjustbox}
}
\caption{\textbf{Ablation study of parameters in oriented bounding box and 24-point polygon representation.} Angle classification and our added IoU loss provide a significant improvement of mAP score relative to a standard baseline. Proposed variable step polygon representation provides significant improvement of 2.7\%. }
\label{tab:ablation}
\end{table}


Compared to standard box representation, oriented box representation is approximately {2.5-4\%} efficient for the side cameras and {1.3-2.3\%} for front cameras. Ellipse improves the efficiency further by an additional 2\% for side cameras and 1-2\% in front cameras.
Our curved box achieves a {1.15\%} improvement over the standard box.
However, it is slightly less than an oriented box due to the constraint that two circular sides of the box share the same circle center, which adds some area inside the polygon, decreasing the IoU. 
In addition, curvature is not modelled for the horizontal edges of the box. In future work, we plan to explore these extensions to obtain a more optimal curved bounding box and leverage the convergence of circular arcs at vanishing points.

The current simple version of curved box has the advantage of getting a tight bottom edge, capturing the footpoint for estimating the object's 3D location. The object's footpoint is captured almost entirely, as observed in qualitative results, especially for the side cameras where distortion is maximized.
Compared to polygon representation, curved-box representation has low annotation cost due to fewer representation points, which saves annotation effort.


\subsubsection{Quantitative Results}

Table \ref{tab:ablation} shows our studies on the methods to predict the orientation of the box or the ellipse efficiently. First, we train a model to regress over the box and its orientation, as specified in equation \eqref{eq:orn_loss}. In the second experiment,  orientation prediction is addressed as a classification problem instead of regression as a possible solution to the discontinuity problem. We divide the orientation range of $180\degree$ into 18 bins, where each bin represents $10\degree$, making this an 18 class classification problem. During the inference, an acceptable error of +-5 degrees for each box is considered. Using this classification strategy, we improve performance by {1.6\%}. We are formulating orientation of box or ellipse prediction as a classification model with IoU loss found to be superior in performance compared to direct regression. This has a {2.9\%} improvement in accuracy. Hence we use this model as a standard representation for oriented box and ellipse prediction when comparing with other representations.\par

\begin{figure*}[t]
  \captionsetup{singlelinecheck=false, font=small, skip=4pt, belowskip=-10pt}
  \centering
  \includegraphics[width=0.98\textwidth]{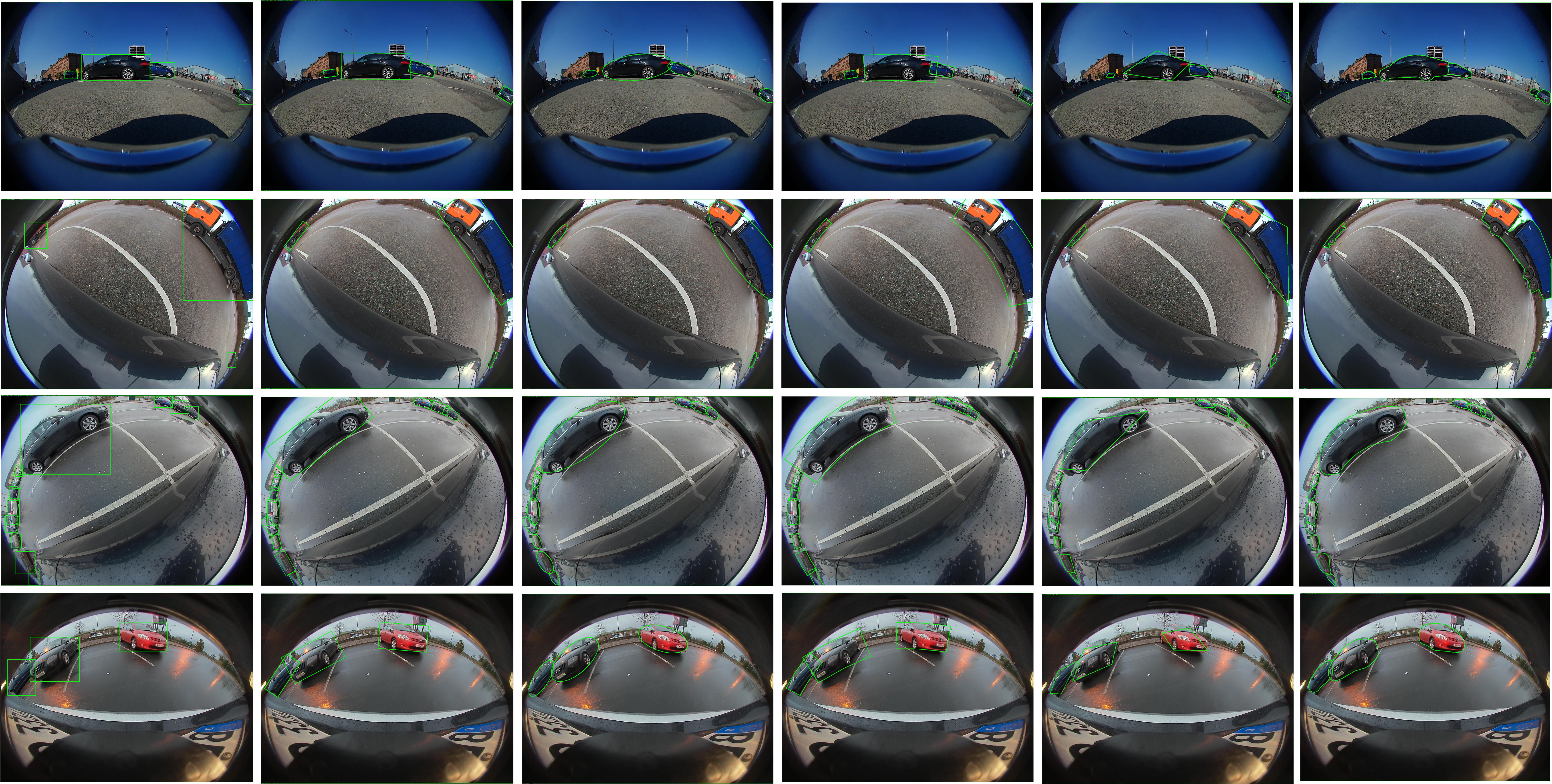}
  \caption{\textbf{Qualitative results of the proposed model outputting different representations.} Rows represent front, left, right, and rear camera images, respectively. \textbf{From left to right:} Standard box, Oriented box, Ellipse, Curved box, 4-point polygon. 24-point polygon.}
\label{fig:qualitative}
\end{figure*}

Table~\ref{tab:prediction_table} demonstrates our prediction results on our proposed representations. Compared to the standard bounding box approach, the proposed oriented box and ellipse models improved mIoU score on the test set by {2\%}, {1.8\%} respectively.
Ellipse prediction provides slightly better accuracy than the oriented box as it has higher immunity to occlusions with other objects in the scene due to the absence of corners, and it is demonstrated in Figure~\ref{fig:qualitative}. 


\subsubsection{Qualitative Results}

\begin{table}[t]
\centering
\captionsetup{belowskip=-12pt, skip=4pt, font= small, singlelinecheck=false}
\begin{adjustbox}{width=0.8 \columnwidth}
\begin{tabular}{l|cccc|c}
\toprule
\multicolumn{1}{l|}{\textit{\textbf{Representation}}} &
  \multicolumn{4}{c|}{\textbf{IoU}} &
  \multicolumn{1}{l}{\textbf{mIoU}} \\\cline{2-5} 
  \multicolumn{1}{l|}{} &
  \multicolumn{1}{l|}{Front} &
  \multicolumn{1}{l|}{Rear} &
  \multicolumn{1}{l|}{Left} &
  \multicolumn{1}{l|}{Right} & \\ \hline
YoloV3       & 32.5 & 32.1 & 34.2 & 27.8 & 31.6 \\
Curved Box & 33 & 32.7 & 35.4 & 28 & 32.3 \\
Oriented Box & 33.9 & 33.5 & 37.2 & 30.1 & 33.6 \\
Ellipse       & 35.4 & 35.4 & 40.4 & 30.5 & 35.4 \\
24-sided Polygon & \textbf{44.4} & \textbf{46.8} & \textbf{44.7} & \textbf{42.7} & \textbf{44.65} \\
\bottomrule
\end{tabular}
\end{adjustbox}
\caption{\textbf{Quantitative results of proposed model on different representations on our dataset.} The experiments are performed on the best performing model according to Table~\ref{tab:gt_table} and Table~\ref{tab:ablation}.}
\label{tab:prediction_table}
\end{table}

Figure~\ref{fig:qualitative} shows a visual evaluation of our proposed representations. Results show that the ellipse provides a decent easy-to-learn representation with a minimum number of parameters and minimum occlusion with the background objects compared to the oriented box representation. Unlike boxes, it allows a minimal representation for the object due to the absence of corners, which avoids incorrect occlusion with free parking slots, for instance, as shown in Figure~\ref{fig:qualitative} (Bottom). Polygon representation provides higher accuracy in terms of IoU with instance mask. A four-point model provides high accuracy predictions with small objects as 4 points are sufficient to represent. As the dataset has significant small objects that helped this representation to demonstrate good accuracy, and the same is shown in Tables \ref{tab:gt_table} and \ref{tab:ablation}. Visually, large objects cannot be represented by a quadrilateral, as illustrated in Figure \ref{fig:qualitative}. A higher number of sampling points on the polygon results in higher performance. However, the predicted masks are still prone to deformation due to minor errors in each point's localization.\par

\section{Conclusion}
{In this paper, we studied various representations of fisheye object detection. At a high level, we can split them into bounding box extensions and generic polygon representations. We explored oriented bounding box, ellipse, and designed a curved bounding box with optimal fisheye distortion properties. 
We proposed a curvature adaptive sampling method for polygon representations, which improves significantly over uniform sampling methods. Overall, the proposed models improve the relative mIoU accuracy significantly by {40\%} compared to a YOLOv3 baseline. We consider our method to be a baseline for further research into this area. We will make the dataset with ground truth annotation for various representations publicly available. We hope this encourages further research in this area leading to mature object detection on raw fisheye imagery.
}
\section*{Acknowledgement}

The authors would like to thank their employer for the opportunity to release a public dataset to encourage more research on fisheye cameras. We also want to thank Lucie Yahiaoui (Valeo) and Ravi Kiran (Navya) for the detailed review.
\bibliographystyle{ieee_fullname}
{\small
\bibliography{bib/egbib}
}
\end{document}